# A Novel Deep Learning Pipeline for Retinal Vessel Detection in Fluorescein Angiography

Li Ding[1], Mohammad H. Bawany[2], Ajay E. Kuriyan[2], Rajeev S. Ramchandran[2], Charles C. Wykoff [3], and Gaurav Sharma[1]

*Abstract*: While recent advances in deep learning have significantly advanced the state of the art for vessel detection in color fundus (CF) images, the success for detecting vessels in fluorescein angiography (FA) has been stymied due to the lack of labeled ground truth datasets. We propose a novel pipeline to detect retinal vessels in FA images using deep neural networks that reduces the effort required for generating labeled ground truth data by combining two key components: cross-modality transfer and human-in-the-loop learning. The cross-modality transfer exploits concurrently captured CF and fundus FA images. Binary vessels maps are first detected from CF images with a pre-trained neural network and then are geometrically registered with and transferred to FA images via robust parametric chamfer alignment to a preliminary FA vessel detection obtained with an unsupervised technique. Using the transferred vessels as initial ground truth labels for deep learning, the human-in-the-loop approach progressively improves the quality of the ground truth labeling by iterating between deep-learning and labeling. The approach significantly reduces manual labeling effort while increasing engagement. We highlight several important considerations for the proposed methodology and validate the performance on three datasets. Experimental results demonstrate that the proposed pipeline significantly reduces the annotation effort and the resulting deep learning methods outperform prior existing FA vessel detection methods by a significant margin. A new public dataset, RECOVERY-FA19, is introduced that includes high-resolution ultra-widefield images and accurately labeled ground truth binary vessel maps.

[1] L. Ding and G. Sharma are with the Department of Electrical and Computer Engineering, University of Rochester, Rochester, NY 14627-0231, USA (e-mail: {l.ding, gaurav.sharma}@rochester.edu).

[2] M. H. Bawany, A. E. Kuriyan, and R. S. Ramchandran are with the University of Rochester Medical Center, University of Rochester, Rochester, NY 14642-0001, USA (e-mail: {mohammad_bawany, ajay_kuriyan, rajeev_ramchandran}@urmc.rochester.edu).

[3] C. C. Wykoff is with Retina Consultants of Houston and Blanton Eye Institute, Houston Methodist Hospital & Weill Cornell Medical College, Houston, TX 77030-2700, USA (e-mail: ccwmd@houstonretina.com).

# 1. Introduction

Recently deep learning based image processing algorithms have shown compelling improvement in the analysis of color fundus (CF) retinal imagery [4], [5], which is the predominant form of retinal images. Deep neural network can detect retinal vessels with high accuracy and robustness [6], [7] and achieve performance close to human experts [8]. Manually labeled ground truth datasets are a key ingredient in the success of these techniques. Three commonly used datasets that provide CF images and corresponding manually labeled pixel-wise binary vessel maps include DRIVE [9] (forty $584 \times 565$ pixel images), STARE [10] (twenty $605 \times 700$ pixel images), and the high resolution HRF [11] (forty-five $3504 \times 2336$ images) datasets. The datasets provide a modest number of images and are used for training in combination with data augmentation techniques [12].

The detection of retinal vessels is also of interest for alternative imaging modalities that are of independent diagnostic utility in the clinic. For instance, fluorescein angiography (FA) and optical coherence tomography angiography (OCT-A) are used for assessing retinal non-perfusion. FA provides a larger field of imaging beyond the macula, while commercially available OCT-A provides more detailed imaging of the macular microvasculature. Although, conceptually, one could redeploy the deep neural network architectures that are successful in CF imagery to these alternative modalities, the fundamental differences between the modalities require fresh training and the lack of ground truth labeled data becomes a key obstacle to such reuse. Specifically, for FA images, only one dataset is available: VAMPIRE [13] which provides eight ultra-widefield FA (UWFFA) images ($3072 \times 3900$ pixels, each) along with limited accuracy ground truth binary vessel maps. Manually annotating vessel maps for training deep neural network is not a trivial task. Specifically, UWFFA images have high resolution and exhibit variations in contrast between the background and the vasculature, which pose a significant challenge for manual annotation. Figure 1 shows sample FA images and highlights the particular challenge of contrast variations. The patch labeled in cyan in the middle UWFFA image is shown in an enlarged view on the right, as captured and with contrast enhanced. From the contrast enhanced view, one can appreciate that the region corresponding to the patch contains a large number of fine vessels that are rather difficult to see without contrast enhancement. In particular, ophthalmologists normally have difficulty in identifying fine vessels in the peripheral region without image enhancement because of the low contrast and brightness. High-quality annotation requires carefully adjusting image contrast for the entire FA image and labeling both major and minor vessels, making it a tedious, time-consuming, and labor-intensive process.

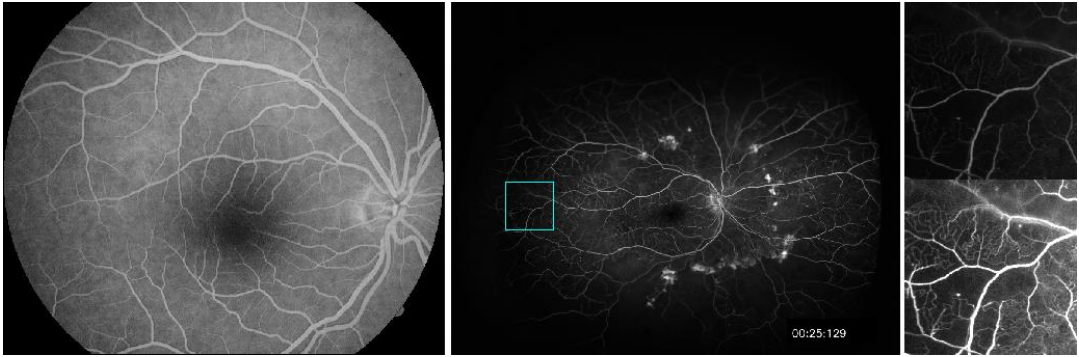

*Figure 1: Sample fluorescein angiography (FA) images. left: fundus FA. Middle: ultra-widefield FA. Right: enlarged view of the cyan rectangle (top and bottom: the original and the contrast-enhanced views, respectively). Compression artifacts in this and subsequent images in the PDF manuscript may impact visual presentation, particularly, for the smaller vessels.*

In this paper, we propose a novel pipeline that enables accurate vessel detection in FA images using deep neural networks by significantly reducing manual annotation effort. The proposed pipeline integrates the following novel elements:

- an unsupervised method for preliminary retinal vessel detection that is based on multiple scales and orientations morphological analysis,

- a cross-modality approach that transfers vessel maps from CF images to FA images using robust chamfer alignment in an Expectation-Maximization (EM) framework, and

- an efficient and effective human-in-the loop iterative deep learning process for detection of retinal vessels in FA imagery that significantly reduces the tedium of generating labeled data.

We demonstrate the utility of the proposed pipeline by developing the first set of deep neural networks for detection of retinal vessels in FA images and evaluating the performance on alternative network architectures. The best performing method provides remarkably accurate results (maximum dice coefficient of 0.854) and offers very significant improvements over the prior methods. Results demonstrate that the approach adapts particularly well to the changes in contrast typical in FA imagery. To facilitate further development of vessel detection in FA images, we also release a new dataset of UWFFA images from the RECOVERY trial [14] along with ground truth labeled vessels from our pipeline. In addition to the innovative pipeline for the generation of training data, demonstration of the first deep learning approaches, evaluation of alternative architectures, and the new ground truth labeled datasets are also contributions of the present work.

The proposed pipeline is also significant from a clinical perspective. FA is a well-established method that provides a useful imaging modality for visualizing, assessing and understanding the impact of diseases on the vascular system. Retinal vasculature changes assessed via FA imagery play a key role in the clinical assessment of vasculature changes caused by multiple common diseases, including diabetes, hypertension, and atherosclerosis, and also for eye-specific diseases, such as retinal venous occlusive diseases and retinal vasculitis. In current clinical practice, ophthalmologists manually review FA images to access disease conditions in retinal vasculature. These examinations are typically qualitative and subjective due to the limited time available during the clinical visits. Quantitative analysis of FA images, although highly desirable, requires inordinate time and patience to be performed manually and thus is not feasible in clinical settings. The proposed pipeline for detecting vessels in FA images offers an automated approach to examine retinal vasculature, which is a key component of computer-assisted retinal image analysis and diagnosis systems. Details of fine vessels are of particular diagnostic significance as changes are often first observed in the fine vessels; a key strength of the method developed is the ability to reliably detect fine vessels, which are often not seen with non-FA modalities and, even for the FA modality, require significant iterative contrast manipulations for visual detection. Using the proposed pipeline, the results of retinal vessel detection achieve a level of accuracy that enables reliable computation of "digital biomarkers" from FA imagery that unlock the potential for improving clinical care, speeding up clinical trials, defining new endpoints of clinical relevance, and characterizing inter-individual variations. Preliminary work demonstrating how the analysis presented here can relate to clinical attributes of interest is being concurrently submitted for a review in a companion paper [15].

The rest of this paper is organized as follows. Section 2 summarizes the existing works on retinal vessel detection. Section 3 provides an overview of the proposed pipeline. In Section 4, we describe the cross-modality transfer for generating ground truth data. In Section 5, we introduce the human-in-the-loop learning approach for both vessel detection and manual annotation. We present the experimental results in Section 6 and summarize concluding remarks in Section 7.

## 2. Related Work

Prior work on detection of vessels in FA imagery is rather limited and due the paucity of ground truth labeled data has been primarily focused on unsupervised techniques. These methods, which are generally rule-based, include hand-crafted matched-filtering [13], active contour models [16], and morphological analysis [1], [17]. The unsupervised methods, however, offer limited accuracy (dice coefficient of 0.634 compared to 0.854 for the best performing method benchmarked here).

Detection of retinal vessels in CF imagery has been extensively studied. For broad context, we refer the readers to a survey [18] and a recent paper [19] that categorize and compare the existing methods. For our discussion, we focus on supervised methods based on deep learning which have significantly advanced the current state of the art for vessel detection in CF images. Various network architectures have been exploited, including per-pixel classifiers [6], [20], fully convolutional networks [8], [21], [22], generative adversarial networks [23], and graphical convolutional network [24]. In addition to the network architectures, several works focus on new loss terms that are particularly attuned to vessel detection [25]–[27]. The basic idea is to incorporate prior knowledge of the topology of vasculature into loss functions.

Recent work in [28] proposes a self-supervised domain adaption work to generate FA images from CF images using a CNN. While this method aims to alleviate the tedium of creating labeled data by utilizing both CF and FA images, the generated pseudo-FA images do not represent actual FA images and normally contain artifacts. In contrast, the proposed pipeline uses a cross-modality approach that directly transfers the vessel map from CF images to FA via robust chamfer registration in an EM framework, and thus is more robust and reliable than the synthesis-based approach.

## 3. Overview of The Proposed Method

The proposed pipeline, illustrated in Fig. 2, has two key components: (1) cross-modality transfer for generating an initial training dataset for FA images from CF images, and (2) a human-in-the-loop learning approach that iteratively refines deep neural networks and expedites the manual annotation process. The cross-modality transfer is inspired by the observation that, in clinical practice, typical retinal imaging protocols capture images with multiple different modalities. For instance, baseline CF images are routinely captured prior to the injection of FA dye. Our proposed pipeline exploits the availability of near concurrently captured CF and FA images in combination with existing deep learning methods for detection of vessels in CF imagery, for which, multiple ground truth annotated datasets are available. The idea is to transfer the ground truth vessel maps from CF images, which are detected using the trained neural network, to FA images. Specifically, we use the publicly available DRIsfahanCFnFA (Diabetic Retinopathy Isfahan Color Fundus and Fluorescein Angiography) dataset [29] ("Unlabeled Joint Dataset in Fig. 2) that contains pairs of CF and FA images that are captured at the same clinical visit but vary in capture viewpoints. A deep neural network (green on in Fig. 2) is trained on existing labeled CF images to extract vessel maps from unlabeled CF images. The detected vessel maps are geometrically aligned with and transferred to FA images via robust chamfer alignment to a preliminary FA vessel maps obtained with morphological analysis [1]. The co-aligned pairs of FA and transformed vessel map ("FA Training Data" in Fig. 2) are used as initial ground truth data to train a deep neural network for vessel detection in FA images.

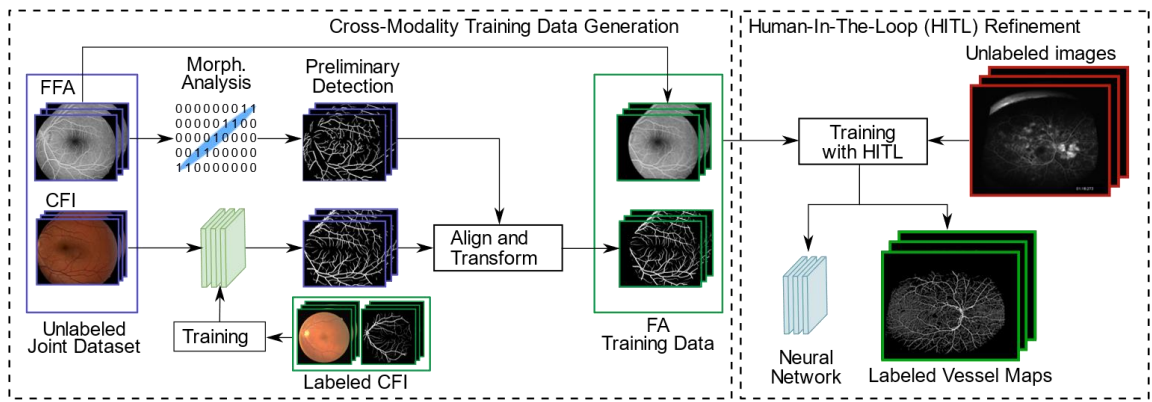

*Figure 2: Overview of the proposed pipeline for vessel detection in FA images. CFI: color fundus images; FFA: fundus fluorescein angiography. The cross-modality transfer (left block) generates the FA training data by aligning vessel maps from CF images with the preliminary vessel maps in FA images. The human-in-the-loop approach (right block) refines the neural network and significantly reduces manual annotation effort.*

The motivation behind the human-in-the-loop learning approach is derived from the synergistic relationship between deep learning and labeling. A well-trained deep neural network model is able to accurately detect vessel maps from FA images. Manually refining the predicted vessel map allows the annotation process to be less time-consuming than labeling the entire image from scratch. On the other hand, enlarging the training dataset with new annotated data is beneficial to further improving the model performance. Therefore, the training and the labeling make each other more effective. We initialize the approach with a deep neural network that is trained on the (approximate ground truth) labeled data generated from the cross-modality transfer. A human annotator then manually refines one or more of the predicted vessel maps to generate improved labeled vessel maps. These vessel maps, manually labeled via the refinement process, are then incorporated into the training dataset to incrementally improve the performance of the deep neural network. We repeat this human-in-the-loop iterative process till the network performance improves significantly and the manual labeling introduces few changes. The end result is a trained deep neural network (shown in blue in Fig. 2) and a set of accurately labeled vessel maps.

Both the cross-modality transfer and the iterative learning approach significantly reduce the burden of manual labeling and thereby engage the annotators more effectively. This engagement is further positively reinforced as the annotator sees the improvement in the trained network performance from iteration to iteration, immediately rewarding them for their effort, instead of requiring a large number of images to be annotated before any improvements are realized. A by product of this engagement and reduction of tedium is that the images are labeled much more accurately than other studies that annotated the images from scratch (see results in Section 6).

# 4. Cross-Modality Ground Truth Transfer

The cross-modality ground truth transfer, illustrated in Fig. 3, generates a training dataset for FA vessel detection from CF images. This approach consists of three steps: (1) vessel detection in CF images using a deep neural network, (2) preliminary vessel detection in FA for anchoring, and (3) vessel registration by parametric chamfer alignment.

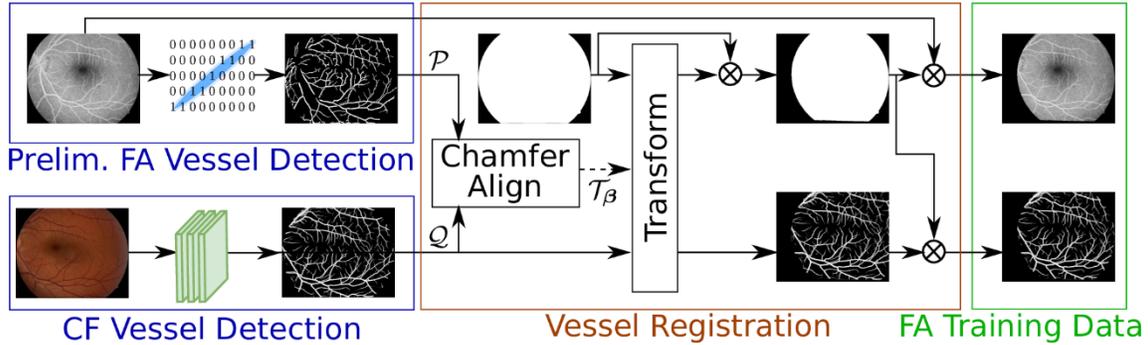

Figure 3: Overview of cross-modality ground truth transfer. The bottom-left shows the vessel detection in unlabeled CF image with neural networks pre-trained on existing CFI dataset. The upper-left shows the preliminary vessel detection in FA obtained with unsupervised morphological analysis. The detected vessels from CF image are transformed to FA via parametric chamfer alignment with vessel maps detected from FA. The overlapping area between CFI and FFA is also estimated. The green block shows the generated training data that includes FA and co-aligned vessel maps that remains in the overlapping area.

## 4.1. Vessel Detection in CF Images

To detect vessels in CF images, we adopt an existing deep neural network proposed in [23] that exploits adversarial learning. The model is trained on DRIVE dataset [9] which scores an Area Under the Receiver Operating Characteristic Curve (ROC AUC) of 0.9803, an Area Under the Precision-Recall curve (ROC PR) of 0.915, and a dice coefficient of 0.829. The pre-trained network is applied to overlapping patches of CF images in the DRIsfahanCFnFA dataset. The final CF binary vessel map is obtained by thresholding the probability map obtained from the generator using Otsu thresholding [30].

## 4.2. Preliminary Vessel Detection In FA Images For Anchoring

A preliminary detection of vessels in FA imagery is obtained using an unsupervised method based on multiple scales and orientations morphological analysis that is attuned to the variations in directions and widths of retinal vessel structure [1]. The input FA image is decomposed into multiple resolutions represented by an image pyramid [31]. Images at each scale are processed independently and the resulting vessel maps at different scales are then combined together to generate a binary vessel map. A Gaussian pyramid expansion is used to resize the vessel maps from each scale to the size of input FA image. Pixels where vessels are detected at any scale are assigned as detected vessels.

The key components in the preliminary vessel detection are the morphological operators that extract structures representing the shape of vessels in the image. Over the retinal vasculature, blood vessels normally have rectilinear structure and are connected in retina. To detect the vessels, we choose a set of linear structuring elements $\mathbf{S}_\alpha$ with the same length but oriented along different angles $\alpha$, ranging from 0° to 180°. We apply the top-hat operator to the FA images using the structuring elements $\mathbf{S}_\alpha$. The conventional top-hat operator [32, p. 557], which is defined as the difference between original and the corresponding morphological opening image, is sensitive to noise. Therefore, we adopt a modified top-hat filtering [33] to improve the robustness of vessel detection. The modified top-hat operator $\odot$ is defined as

$$\mathbf{X} \odot \mathbf{S}_\alpha = \mathbf{X} - min((\mathbf{X} \bullet \mathbf{S}_\alpha) \circ \mathbf{S}_\alpha, \mathbf{X}), \tag{1}$$

where $\mathbf{X}$ is the input image, and $\bullet$ and $\circ$ indicate the morphological operators of image closing and opening, respectively.

The morphological operation yields an image where pixel intensities correspond to the responses from the modified top-hat operation. Pixels with high response values are likely to be vessels whereas pixels with low intensities are more likely to be background. We convert this soft vessel segmentation into binary vasculature map with the adaptive thresholding method. The binary vessel image obtained has a few disconnected components. As a post-processing step, we perform an area opening operation to remove all small segments from the vessel map.

## 4.3. Vessel Registration By Chamfer Alignment

To precisely transfer the vessel maps in CF images to the corresponding FA images, we use parametric chamfer alignment in an EM framework [2]. Let $\mathcal{P} = \{\mathbf{p}_i\}_{i=1}^{N_i}$ and $\mathcal{Q} = \{\mathbf{q}_j\}_{j=1}^{N_j}$ be two sets of reference and targets points corresponding to the coordinates of the vessel pixels in FA and CF images, respectively, where $\mathbf{p}_i = (x_i, y_i)^\top$ and $\mathbf{q}_j = (u_j, v_j)^\top$. We adopt a second-order polynomial transformation to align the two sets of coordinate vectors for points corresponding to detected vessels. Specifically, the coordinate vector $\mathbf{q}_j$ for the $j^{\text{th}}$ point is mapped to the coordinate vector

$$\mathcal{T}_\boldsymbol{\beta}(\mathbf{q}_j) = \begin{bmatrix} \beta_1 \\ \beta_7 \end{bmatrix} + \begin{bmatrix} \beta_2 & \beta_3 \\ \beta_8 & \beta_9 \end{bmatrix} \begin{bmatrix} u_j \\ v_j \end{bmatrix} + \begin{bmatrix} \beta_4 & \beta_5 & \beta_6 \\ \beta_{10} & \beta_{11} & \beta_{12} \end{bmatrix} \begin{bmatrix} u_j^2 \\ u_j v_j \\ v_j^2 \end{bmatrix}, \tag{2}$$

where $\boldsymbol{\beta} = \{\beta_i\}_{i=1}^{12}$ are the transformation parameters and $\mathcal{T}_\boldsymbol{\beta}$ denotes the geometric transformation. The alignment error $d_j(\boldsymbol{\beta})$ for the $j^{\text{th}}$ point under the geometric transformation $\mathcal{T}_\boldsymbol{\beta}$, is quantified as the minimum squared Euclidean distance between the transformed location $\mathcal{T}_\boldsymbol{\beta}(\mathbf{q}_j)$ and the nearest point from $\mathcal{P}$, viz.,

$$d_j(\boldsymbol{\beta}) = \min_i \| \mathbf{p}_i - \mathcal{T}_{\boldsymbol{\beta}}(\mathbf{q}_j) \|^2. \tag{3}$$

In the absence of outliers, the parameters $\boldsymbol{\beta}$ can be estimated by minimizing the average of the errors in (3), which corresponds to conventional chamfer minimization [34]. The method is, however, sensitive to outliers, that are inevitable in the detection process due to stochastic variations and noise in the imaging processes and due to differences in the FOV between the modalities. Particularly, vessel pixels in $\mathcal{Q}$ that do not have corresponding points in $\mathcal{P}$ inevitably cause the chamfer minimization to converge to a poor local minima, resulting in poor registration. To tackle this issue, we adopt a probabilistic formulation of chamfer alignment in an EM framework. Specifically, we introduce latent binary variables $W_j \in \{0,1\}$ to assess putative correspondence between vessel pixel $\mathbf{q}_j$ in CF images and vessel pixels $\mathcal{P}$ in FA, where $W_j = 1$ indicates that $\mathbf{q}_j$ has corresponding points in $\mathcal{P}$ and thus is not an outlier point, and $W_j = 0$ otherwise. The prior probability of latent variable $W_j$ follows a Bernoulli distribution with parameter $\pi = p(W_j = 1)$. Under the assumption that the points correspond, the transformed inlier vessel pixels in CF image should be located in close proximity to the vessel pixels in FA. Therefore, the alignment error is modeled is modeled as an exponential distribution with parameter $\lambda$. For outlier points, we model the alignment error as an uniform distribution over the interval $[0, D_{max}]$, where $D_{max}$ is a free parameter. Specifically, conditioned on the latent variable and the parameters $\boldsymbol{\theta} = \{\pi, \lambda, \boldsymbol{\beta}\}$, the distribution of the random variable $D_j$ corresponding to the squared distance in (3) is modeled as

$$p_{D_j|W_j,\boldsymbol{\theta}}(d_j|w_j, \boldsymbol{\theta}) = \begin{cases} \lambda e^{-\lambda d_j}, & \text{if } w_j = 1 \\ \dfrac{1}{D_{max}}, & \text{if } w_j = 0 \end{cases} \tag{4}$$

The EM algorithm seeks to obtain a maximum likelihood estimate of the parameters $\boldsymbol{\theta}$ via an iterative procedure comprising two steps: an expectation (E) step and a maximization (M) step. At the $(l+1)^{\text{th}}$ iteration, the E-step computes the expectation $Q(\boldsymbol{\theta}, \widehat{\boldsymbol{\theta}}^{(l)})$ of the complete-data log-likelihood

$$L_c(\boldsymbol{\theta}) = \sum_{j=1}^{N_j} \log p(d_j, w_j | \boldsymbol{\theta}), \tag{5}$$

given the current estimate $\widehat{\boldsymbol{\theta}}^{(l)}$ of the parameters. In the M-step, the updated parameters $\widehat{\boldsymbol{\theta}}^{(l+1)}$ are determined by maximizing $Q(\boldsymbol{\theta}, \widehat{\boldsymbol{\theta}}^{(l)})$. For our specific setting, the E-step reduces to a computation of the posterior probabilities $p_j^{(l)} = p(W_j = 1|d_j, \boldsymbol{\theta}^{(l)})$, which are obtained as

$$p_j^{(l)} = \frac{\pi^{(l)}\lambda^{(l)}e^{-\lambda^{(l)}d_j}}{\pi^{(l)}\lambda^{(l)}e^{-\lambda^{(l)}d_j}+(1-\pi^{(l)})\frac{1}{D_{max}}}, \tag{6}$$

The updates in the M-step become

$$\widehat{\pi}^{(l+1)} = \frac{\sum_{j=1}^{N_j} p_j^{(l)}}{N_j}, \qquad \widehat{\lambda}^{(l+1)} = \frac{\sum_{j=1}^{N_j} p_j^{(l)}}{\sum_{j=1}^{N_j} p_j^{(l)} d_j}, \tag{7}$$

with the updated transformation parameter $\boldsymbol{\beta}^{(l+1)}$ given by

$$\widehat{\boldsymbol{\beta}}^{(l+1)} = \arg\min_{\boldsymbol{\beta}} \frac{1}{N_j} \sum_{j=1}^{N_j} p_j^{(l)} d_j(\boldsymbol{\beta}). \tag{8}$$

By examining (8), we see that the optimal parameters are obtained by minimizing the weighted average chamfer distance where the weighting for each datapoint equals the posterior probability that it is not an outlier. This makes intuitive sense, with the EM framework, the weighting concentrates on non-outliers and discounts the impact of outliers, making the parameter estimates much more robust than direct (non-probabilistic) chamfer minimization.

The optimization problem in (8) can be solved using the iterative Levenberg-Marquardt (LM) non-linear least squares algorithm [35] in combination with suitable distance transforms [36] that significantly simplify the computation of the objective function and required gradients with respect to the parameters $\boldsymbol{\beta}$. Detailed derivations of parameter update equations listed above are provided in Section S.2 in the supplementary material.

The LM algorithm based transformation parameter updates in (8) can get trapped in poor local minima. This is because the LM algorithm strongly depends on the initial parameter $\widehat{\boldsymbol{\beta}}^{(0)}$. Thus, a good initialization is important to obtain a good solution. Instead of estimating all 12 parameters from scratch, the optimization in (8) is further performed in progressive steps that use Euclidean, similarity, affine, projective (homography), and second-order polynomial transformations, in sequence. The EM iterations are terminated when the changes in the updates become smaller than a tolerance threshold and the final estimates $\widehat{\boldsymbol{\beta}}$ for the transformation parameters are set to the values from the last iteration.

The binary vessel maps in CF images are registered to the corresponding FA images by applying the transformation $\mathcal{T}_{\widehat{\boldsymbol{\beta}}}$. To select common region where retina surface is captured in both CF and FA images, we first generate a binary mask for original CF, which is then transformed using the same transformation used for the binary vessel map. The mask for the overlapping area can be readily obtained as the intersection of the transformed mask and the original one. Only pixels remaining in the common area are selected as the inferred training data for initiating the next stage of the pipeline.

Parametric chamfer alignment is an ideal tool for registering images from different modalities. First, given the asymmetry of the chamfer distance, the preliminary vessel detector can be chosen to have a high specificity but a relatively low sensitivity. This means that the results of preliminary vessel detection have a low false positive rate, even though the corresponding true positive rate is low as well. In addition, the formulation uses a global matching of the detected vessels rather than relying on a set of key points with feature descriptors, which is beneficial for the polynomial parametric mapping. Finally, the incorporation of EM framework for parameter estimation significantly enhances the robustness of the registration by mitigating the effects of outlier vessel points.

As a method for generating training data for FA vessel detection, the proposed cross-modality transfer has the benefit of contrast invariance because the inferred vessels are transformed from those detected in CF images. Figure 4a and 4b show two FA images in DRIsfahanCFnFA dataset with significant variation in contrast. The corresponding vessel maps, which are shown in Fig. 4b and 4d, respectively, provide consistent detection, regardless of image contrast, and capture both major and minor vessels.

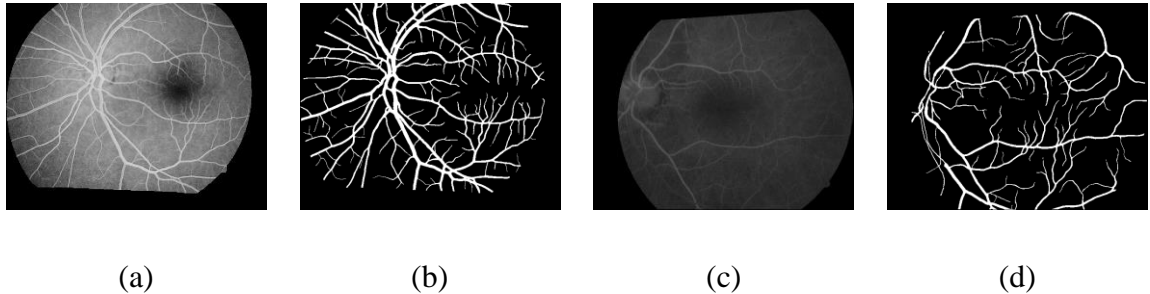

(a) (b) (c) (d)

*Figure 4: Sample results of generated training data for FA imagery in DRIsfahanCFnFA dataset. (a) and (c) show two FA images, and (b) and (d) are the corresponding vessel maps. Notice that the generated vessel maps are robust under different contrast conditions.*

## 5. Human-in-the-loop Iterative Learning/Labeling

Although the cross-modality transfer allows generation of a reasonable labeled dataset for training deep neural networks for detecting vessels in FA images, the accuracy of the labeling is limited by the differences between the modalities and the performance limitations of the CF vessel detection. The network performance can be significantly improved by providing additional better labeled ground truth data.

As indicated in Section 1, manually annotating a high-resolution UWFFA image is particularly tedious and time-consuming. In this section, we present the human-in-the-loop learning approach that aims to further refine the deep neural network by incorporating more training data and to facilitate and expedite the manual annotation process. Figure 5 contrasts the conventional approach to annotation of training data against the proposed human-in-the-loop approach. For conventional approach, the annotation and the training are carried out in separate sequential phases, meaning that all images in the dataset are first annotated and then used for the training stage. The human-in-the-loop approach, however, is an iterative process that exploits the synergistic relationship between deep learning and labeling. The process is initialized with a trained deep neural network trained to detect vessels in FA images using the training data obtained by the cross-modality transfer approach of Section 4. Estimated binary vessel maps that indicate the pixels corresponding to vessels are obtained for a small subset of images from an unlabeled (FA-only) dataset and used as the as the starting point for manual annotation. Specifically, the human annotator corrects the estimated binary vessel map by removing false positive detections and adding in false negative detections. The new labeled images are incorporated into the training dataset to refine the deep neural network in the next iteration. This process is repeated until all images are labeled.

The proposed human-in-the-loop approach radically reduces the effort required for annotating images (see the discussion in Section 6.2 where the experiments are described). In addition to reducing the time and tedium for annotation, the approach also benefits from a psychological advantage that it provides. The annotators see the improvements in the trained network from iteration to iteration and feel immediately rewarded for their effort instead of having to label many images before seeing any machine generated annotations. This engages annotators much better than *de novo* labeling approaches, analogous to how gamification of learning and education generates better engagement [37], [38]. Our results indicate that the approach generates significantly better labeled data than the traditional *de novo* labeling approach.

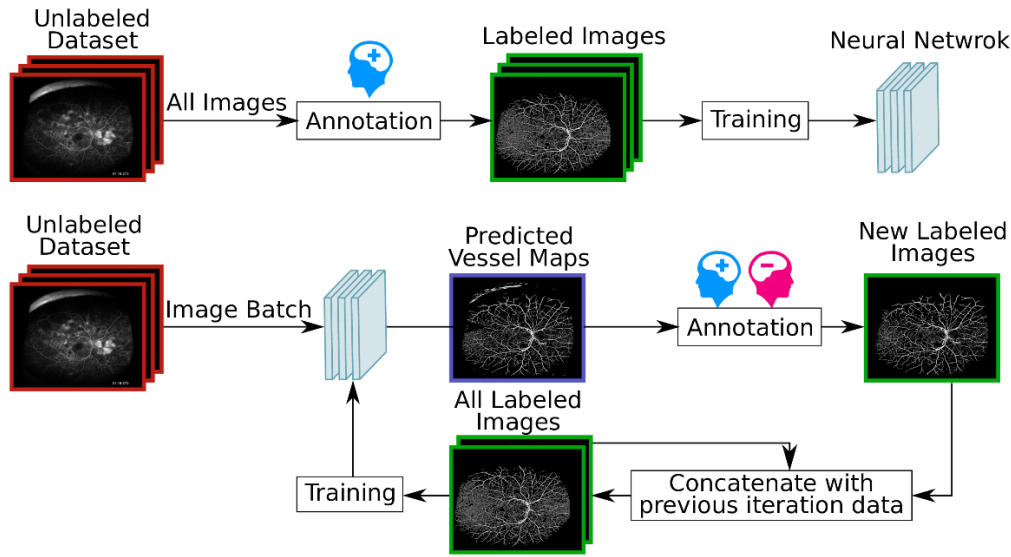

*Figure 5: Annotation and training pipelines. Top: conventional approach starts with manual annotation that generates ground truth for all images and then trains neural network with the ground truth data. Bottom: the proposed human-in-the-loop approach iterates between training neural network and manually correcting annotations generated for a batch of images using a trained network from the previous iteration.*

## 5.1. Network Architecture

We trained and evaluated a number of alternative deep neural network architectures for vessel detection in FA images. In this section, we describe the best performing approach that exploits the recent concept of generative adversarial network (GAN) [39], which was also the architecture used for the human-in-the loop labeling iterations. Detailed architectures for other neural networks are provided in Section S.3 in the supplementary material. To apply GAN to vessel detection, we formulate the problem as an image-to-image translation [40]. In this context, the network consists of a generator $\mathcal{G}$, which is trained to learn a mapping from the FA image **X** to the vessel map **V**, and a discriminator $\mathcal{D}$, which aims to distinguish between real pairs $(\mathbf{X}, \mathbf{V})$ and generated pairs $(\mathbf{X}, \mathcal{G}(\mathbf{X}))$ of FA images and vessel maps. The idea is to jointly train $\mathcal{G}$ and $\mathcal{D}$ to achieve the min max operating point where the vessel maps generated by $\mathcal{G}$ minimize the maximum error for the discriminator $\mathcal{D}$ in distinguishing between real and generated pairs.

The network architecture is visualized in Fig. 6. For the generator, we adopt the UNet [41] architecture, which comprises a downsampling path and an upsampling path. The key component in the UNet is the skip-connection that concatenates each upsampled feature map with the corresponding one in the downsampling path that has the same spatial resolution. The skip-connection is designed for detecting fine vessel structures. The discriminator receives either an image pair $(\mathbf{X}, \mathbf{V})$ (the blue and green bars) or $(\mathbf{X}, \mathcal{G}(\mathbf{X}))$ (the blue and yellow bars).

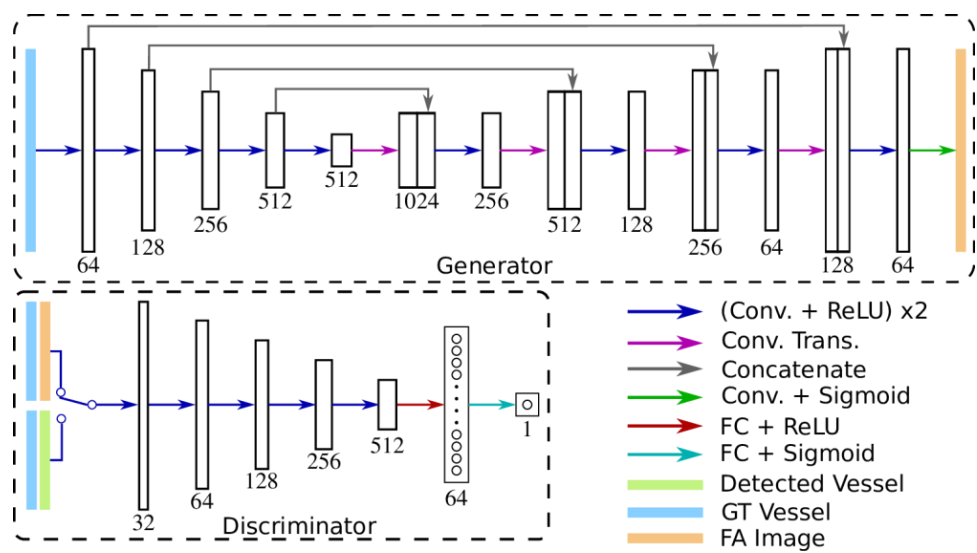

*Figure 6: Network architecture for the GAN network used with the proposed pipeline. The rectangular blocks are feature maps where heights indicate spatial dimensions. The last two blocks in the discriminator show the outputs from fully connected layers. The numbers below the rectangular block show the number of feature channels (or number of hidden units for fully connected layers).*

## 5.2. Training

The objective function for the GAN is defined as

$$\mathcal{L}_{GAN} = \mathbb{E}_{\mathbf{X},\mathbf{V}}[\log\mathcal{D}(\mathbf{X},\mathbf{V})] + \mathbb{E}_{\mathbf{X}}[\log(1 - \mathcal{D}(\mathbf{X},\mathcal{G}(\mathbf{X})))], \qquad (9)$$

where minimization of the first and the second terms encourage correct classification by the discriminator $\mathcal{D}$ of real pairs $(\mathbf{X}, \mathbf{V})$ sampled from training set and the pairs $(\mathbf{X}, \mathcal{G}(\mathbf{X}))$ generated by $\mathcal{G}$, respectively.

Inspired by the idea proposed in [40] that integrates a data loss ($\ell_1$ loss) into the objective function, we combine the objective function in (9) with the binary-cross entropy loss commonly used for segmentation. Specifically, we use the segmentation loss

$$\mathcal{L}_s = -\mathbb{E}_{\mathbf{X},\mathbf{V}}[\mathbf{V}\log\mathcal{G}(\mathbf{X}) + (1 - \mathbf{V})\log(1 - \mathcal{G}(\mathbf{X}))], \qquad (10)$$

which penalizes the disagreement between the estimated and ground truth vessel map. The training procedure is then a min-max game [39] between the generator and the discriminator

$$\min_{\mathcal{G}}\max_{\mathcal{D}} \mathcal{L}_{GAN}(\mathcal{G}, \mathcal{D}) + \lambda \mathcal{L}_s(\mathcal{G}), \qquad (11)$$

where $\lambda$ is the free parameter to control the relation between GAN loss and segmentation loss. The trained deep network $\mathcal{G}$ obtained from this procedure is used to detect vessels in FA images.

## 6. Experiments

We begin by summarizing the implementation parameters, listing alternative vessel detection methods that we use as baselines for comparison, and defining the evaluation metrics that we use. We then structure our presentation of the results as follows. First, we highlight the operation and benefit of the proposed pipeline, illustrating how the cross-modality transfer and the human-in-the-loop approach reduce the burden of annotation and yield our accurately labeled RECOVERY-FA19 dataset. Next, we evaluate the performance of alternative network architectures on the UWFFA RECOVERY-FA19 dataset, using a cross-validation approach that ensures that the test datasets are independent of the training datasets. Additionally, we demonstrate the broader utility of the trained networks for vessel detection in FA images, by quantifying the performance on the VAMPIRE [13] dataset and the DRIsfahanCFnFA [29] dataset, the first of which consists of UWFFA images from a source that is entirely independent of the RECOVERY-FA19 dataset and the second of fundus FA images.

### 6.1. Implementation, Baselines, and Evaluation Metrics

The preliminary vessel detection and chamfer registration discussed in Section 4 are implemented in MATLAB. Using the training data generated with the proposed pipeline, we assess the performance of several alternative deep neural network architectures for FA vessel detection. Specifically, we use the UNet [41], NestUNet [42], and GAN [39] architectures, where, as described in Section 5 the GAN uses UNet [41] as the generator. The deep neural networks are implemented using PyTorch [43] (Version 0.4.1). Detailed parameter settings and training protocol are provided in Section S.3 of the supplementary material. As baselines for performance comparisons, we use the following existing methods for vessel detection in FA images: SFAT [13], MSMA [1], and VDGAN [3].

For quantitative comparison, we use the Receiver Operating Characteristic (ROC) curve and the Precision-Recall (PR) curve. The ROC curve is plotted with the true positive rate (TPR, or recall) against the false positive rate (FPR) at multiple threshold values between 0 and 1, and the PR curve is a plot of the precision versus the recall for various thresholds. We also report the area under curve (AUC) and the maximum dice coefficient (DC, or F1 score) as summary measures. These metrics can be computed from the true positive (TP), false positive (FP), true negative (TN), and false negative (FN) as

$$\text{Recall} = \frac{TP}{TP+FN} \quad FPR = \frac{FP}{FP+TN}$$
$$\text{Precision} = \frac{TP}{TP+FP} \quad DC = \frac{2TP}{2TP+FP+FN}.$$

## 6.2. Annotation of the RECOVERY-FA19 Dataset

Images for the RECOVERY-FA19 dataset were selected from those gathered for the Intravitreal Aflibercept for Retinal Non-Perfusion in Proliferative Diabetic Retinopathy trial (RECOVERY, ClinicalTrials.gov Identified: NCT02863354) [14]. The dataset comprises eight high resolution ($3900 \times 3072$ pixels) UWFFA images in 8-bit TIFF format acquired using Optos California and 200Tx cameras with a 200° FOV of the retina [44]. Ground truth binary vessel map annotations were obtained for the images using the proposed pipeline described in Sections 3-5. In each human-in-the loop iteration, the network-predicted vessel map was refined by an annotator. The refinement annotations were performed using the Fiji distribution of ImageJ [45] with the segmentation editor plugin, which allows the (current estimate of the) vessel map to be overlaid on the UWFFA image to facilitate annotation. The brush tool, polygon selection, and freehand selection tools available in Fiji were used to add and remove pixels in the vessel map. The annotator adjusted the brightness and contrast of the UWFFA images to accurately identify the vessels. The annotations were validated by consultation with two ophthalmologists who routinely use UWFFA images for diagnosis in their clinical practice and research.

To validate that the proposed pipeline can reduce the burden of annotation, at each iteration, we calculate the number of pixels changed from the network-predicted vessel map in the human-annotation process. Table 1 lists the number of pixels added and removed during the iterative annotation process for seven iterations. The traditional *de novo* labeling approach on average requires annotation of an estimated $1.1M$ pixels in each image. Using the proposed pipeline, in the first iteration, $36.6\%(292.4K)$ pixels were added and $0.87\%(6.9K)$ pixels were removed from the initial vessel map generated from the training data obtained using the cross-modality transfer approach of Section 4, which is *very significantly reduced compared with labeling from scratch*. This highlights the benefit of the cross-modality transfer approach, while also illustrating the need for improvement beyond what is achieved with that approach. Specifically, the FOV for the CF modality is smaller than for UWFFA and therefore the training dataset generated with the cross-modality transfer approach lacks fine vessel structure seen in the peripheral regions for UWFFA. As a result, in the first iteration the annotator added a significant number of pixels. As the human-in-the-loop iterations proceed, and newly labeled images are incorporated into the training dataset, the performance of deep neural network progressively improves requiring fewer and fewer annotation changes. In the last ($7^{th}$) iteration, only $2\%(19.3K)$ pixels are added and only $1.4\%(14.1K)$ pixels are removed. As noted in Section 5, the progressive improvements in the network performance also have a positive psychological impact as the annotator realizes that the tedium of labeling is progressively decreasing.

| Iteration | # images | # pixels added | # pixel removed |
|---|---|---|---|
| 0* | - | 1.1M (100%) | 0.0 |
| 1 | 1 | 292.4K (36.6 %) | 6.9K (0.87 %) |
| 2 | 2 | 79.1K (13.0 %) | 13.0K (0.99 %) |
| 3 | 1 | 42.1K (3.8 %) | 7.8K (0.7 %) |
| 4 | 1 | 32.7K (2.9%) | 14.1K (1.3%) |
| 5 | 1 | 21.4K (1.7%) | 9.1K (0.7%) |
| 6 | 1 | 20.4K (1.5%) | 26.2K (1.9%) |
| 7 | 1 | 19:3K (2.0%) | 14:1K (1.4%) |

*Table 1: Number of pixels changed in each iteration in the proposed human-in-the-loop process. * The row labeled iteration 0 lists the estimated number of pixels that would need to be added to a vessel map, starting from scratch.*

The annotated vessel maps obtained by the human-in-the-loop iterations along with the corresponding eight UWFFA images constitute a new labeled dataset for vessel detection in FA images, which we refer to as the RECOVERY-FA19 dataset [46]. The RECOVERY-FA19 dataset contains fine vessel branches, leakage, neo-vasculation, and retinal non-perfusion, which make the vessel detection more challenging. These attributes are of particular diagnostic significance [47] but are barely seen in the prior VAMPIRE dataset [13]. Figure 7 shows an example of labeled ground truth vessel map for the UWFFA image in Fig. 1. The ground truth annotations for RECOVERY-FA19 are also significantly better than for VAMPIRE, which we attribute primarily to the pipeline proposed in this paper, which significantly reduces the tedium of labeling and significantly improves annotator engagement.

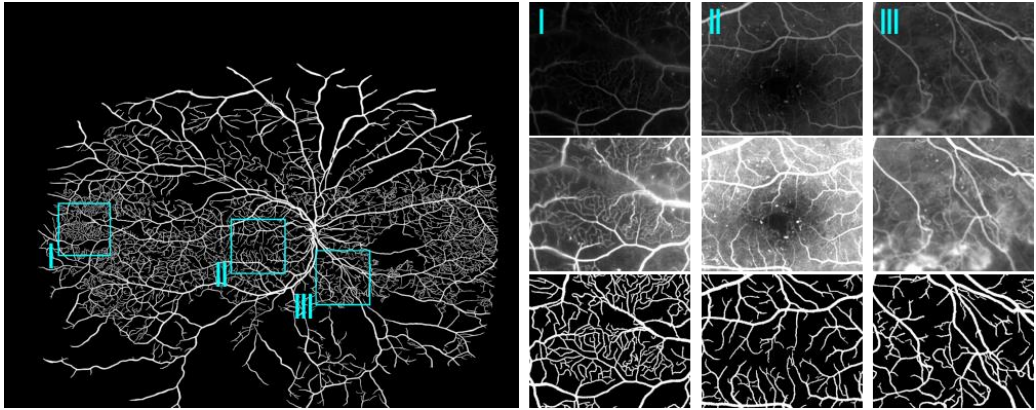

*Figure 7: Example of labeled ground truth vessel map from the RECOVERY-FA19 dataset. Enlarged views of cyan rectangles are shown on the right (top: original view; middle: contrast-enhanced view; bottom: labeled ground truth). The corresponding UWFFA image is shown in Fig. 1.*

## 6.3. Evaluations on the RECOVERY-FA19 Dataset

In the course of the human-in-the-loop iterations, labeled ground truth data is co-mingled with prior iteration training data to generate training data for the next iteration. A limitation of this setting is that only the added ground truth data at each iteration is the "test" data independent of the training data. Therefore, in order to evaluate the performance of alternative network architectures on the RECOVERY-FA19 dataset we use the standard leave-one-out cross validation [48] process commonly used with small ground truth datasets. The model is trained on seven of the eight UWFFA images and the corresponding ground truth vessel map labels and tested on the remaining image for evaluation. This process is repeated eight times, where each time a different image is left out. The performance of the model is then reported in terms of statistics of the evaluation metrics of the eight evaluations. The approach ensures that each labeled image contributes to the performance evaluation while still ensuring that the test set is completely independent of the training data.

Figure 9 shows the ROC and the PR curves for different methods and Table 2 summarizes the AUC for both curves and the maximum DC. The best performing network (Prop. + GAN) achieves an AUC ROC of 0.987, an AUC PR of 0.930, and the maximum DC of 0.854. Using the proposed pipeline, all deep neural networks show significant improvement over traditional methods SFAT [13] and MSMA [1]. The performance is also significantly better than that obtained with the precursor to the present work [3], where only the cross-modality transfer was used. This highlights the benefit of the human-in-the loop iterations in the proposed pipeline. In Figure 8, we show qualitative results of different methods. Notice that the proposed pipeline is robust to contrast variations. Fine vessels are detected in the periphery that has extremely low contrast and brightness. Although these details in vasculature can be seen manually by repeatedly adjusting contrast and viewing different regions, the burden and the time requirement for doing this are prohibitive in typical clinical settings. The proposed pipeline also handles capillary leakage, neo-vasculation, and retinal non-perfusion, as shown in the enlarged views in Fig. 8.

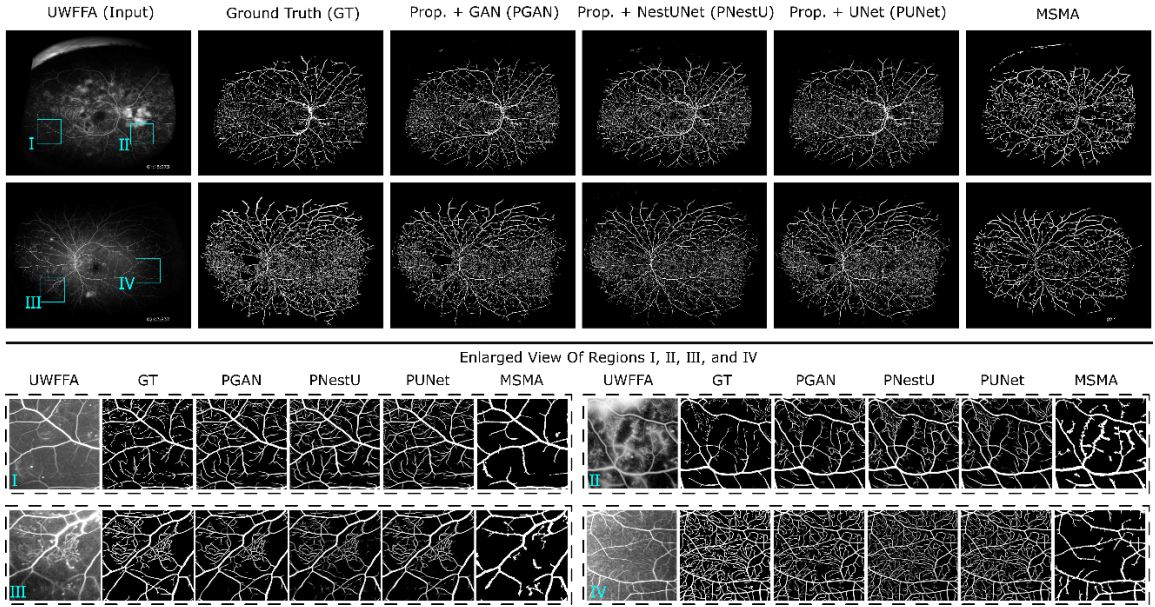

Figure 8: Qualitative comparison of results obtained with different algorithms for images from the RECOVERY-FA19 dataset. For each full image, two contrast-enhanced enlarged views of the selected regions (shown by cyan rectangles) are also included.

| Methods | AUC ROC | AUC PR | Max DC |
|---:|---:|---:|---:|
| SFAT [13] | - | - | 0.606 |
| MSMA [1] | - | - | 0.634 |
| VDGAN [3] | 0.981 | 0.883 | 0.800 |
| Prop. + UNet | 0.987 | 0.923 | 0.842 |
| Prop. + NestUNet | 0.955 | 0.900 | 0.817 |
| Prop. + GAN | **0.987** | **0.930** | **0.854** |

Table 2: Quantitative comparison of the results obtained from different methods on the RECOVERY-FA19 dataset. The best result is shown in bold.

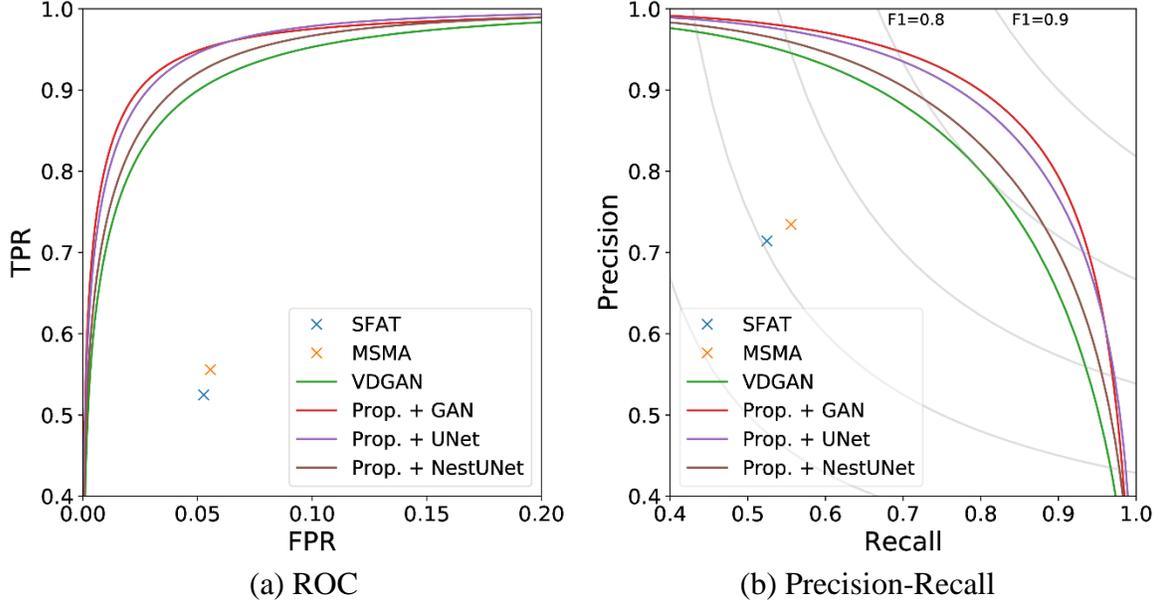

(a) ROC  (b) Precision-Recall

Figure 9: *ROC and Precision-Recall curves for different methods on the RECOVERY-FA19 dataset. The gray curves represent the isolines of dice coefficients.*

## 6.4. Evaluations on the VAMPIRE Dataset

The VAMPIRE dataset [13] provides eight high resolution ($3900 \times 3072$ pixels) UWFFA images acquired using the OPTOS P200C camera [44] with a 200° FOV of the retina. There are two sequences of images in the VAMPIRE dataset representing a healthy retina (GER) and a retina with age-related macular degeneration (AMD). For each image, a binary vessel map that is manually annotated by ophthalmologists is provided as ground truth.

We detected vessels in the UWFFA images from the VAMPIRE dataset using the best performing (Prop+GAN) network that was trained on the RECOVERY-FA19 dataset. Our results reveal an issue with the VAMPIRE dataset: we notice that the vessels branches are not fully-labeled, especially in peripheral regions where the images have extremely low contrast. As mentioned in Section 1, contrast and exposure normally pose a big challenge for manual annotation. To demonstrate the issue, we visually examine the result for the image "AMD2" in the VAMPIRE dataset, as shown in Fig. 10. Using the labeled vessel map provided with the VAMPIRE dataset as "ground truth", we visualize true positive (black), false positive (red), false negative (blue), and true negative (white), as shown in the middle image in the first row of Fig. 10. After closely examining the vessel detection results, we observe that most "false positive" detections are indeed true vessels but are not annotated in the original labeling. For example, the second and the third rows of Fig. 10 show six rectangular regions where the true vessel branches are missed. This illustrates that quantitative comparisons using the original labeling for the VAMPIRE dataset are not reliable. To remedy the situation, we selected two images,

"AMD2" and "GER4", from the dataset and obtained (refined) ground truth vessel map annotations for these using the human-in-the-loop approach. The fourth row of Fig. 10 shows the same enlarged views as earlier, evaluated on the ground truth images. Compared with the evaluation using the original labeling (the third row of Fig. 10), the evaluation using the ground truth data indicates that the detected vessel map has much less false positives. On average, 73% of the original false positive detections becomes true positive if they are evaluated using the ground truth.

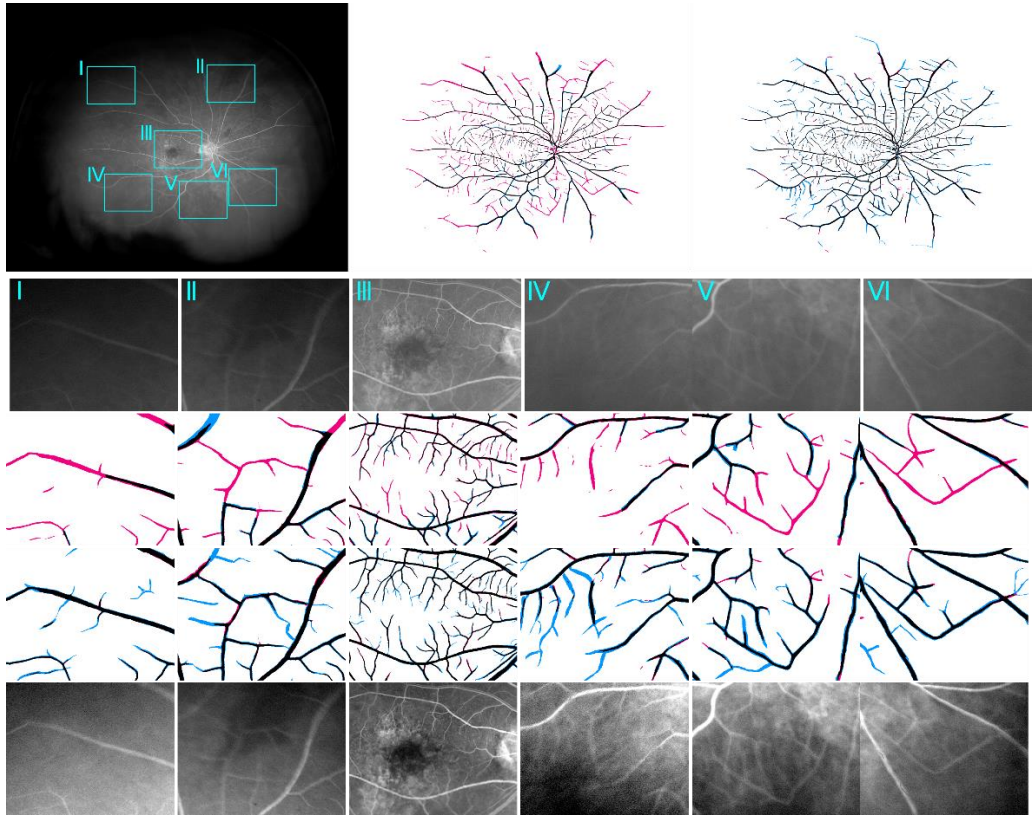

*Figure 10: Sample results of vessel detection on the VAMPIRE dataset [13]. The first row, from left to right: UWFFA, vessel map evaluated on the original VAMPIRE ground truth, and the vessel map evaluated on the refined ground truth. Black, red, and blue indicates true positive, false positive, and false negative, respectively. The second to the fourth rows show the enlarged views of six rectangular regions marked on the wide-filed FA images and corresponding results, respectively. The "false positive" detections in the third row are actually true vessels that are not labeled in the VAMPIRE dataset. In the last row, we show the images after contrast enhancement for a better visualization.*

ROC and the PR curves in Fig. 11 present quantitative evaluations of alternative vessel methods using the (refined) ground truth. The Prop. + GAN network achieves an AUC ROC of 0.995, an AUC PR of 0.50, and maximum DC of 0.878. For reference, we also plot the accuracy of the original (vessel) labeling on the plots (shown by the points marked by the blue crosses ×). These plots highlight the fact that the network-predicted vessel maps are significantly better than original vessel map labels, which further demonstrates the utility of the proposed pipeline. For completeness, in Section S.3.3 of the supplementary material, we also provide quantitative evaluations performed with the original labeling and contrast these against evaluations over the two images with refined ground truth.

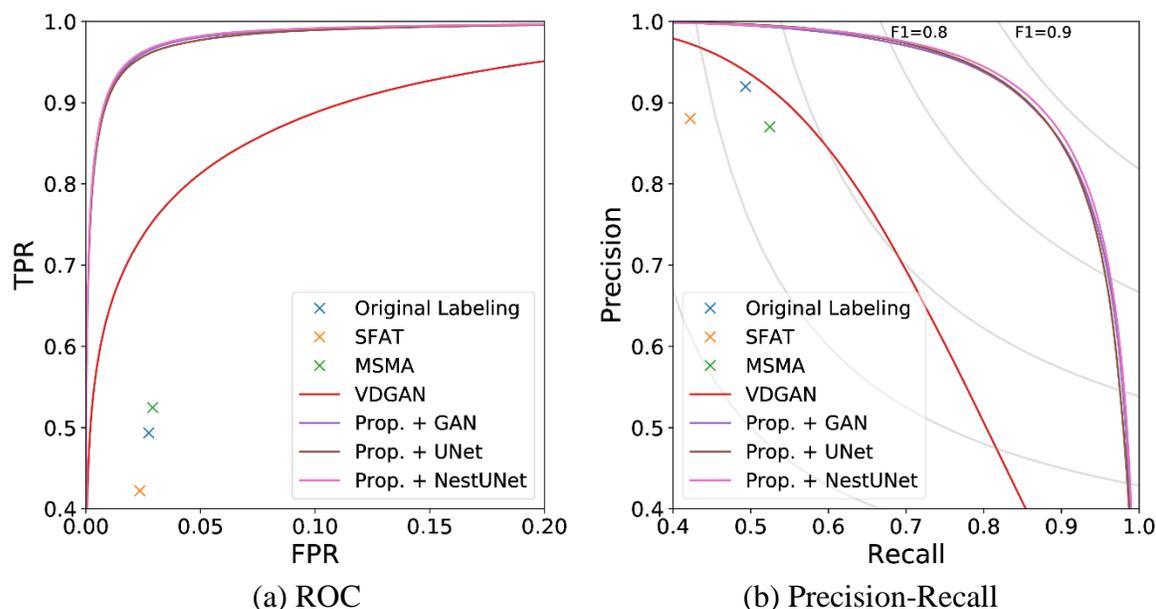

(a) ROC  (b) Precision-Recall

Figure 11: ROC and Precision-Recall curves for different methods. The results are evaluated on the "AMD2" and the "GER4" images from the VAMPIRE [13] dataset with the refined ground truth. The gray curves represent the isolines of dice coefficients.

## 6.5. Evaluations on the DRIsfahanCFnFA Dataset

The FA imaging modality shares common physical characteristics across alternative imaging options and therefore the proposed methodology is useful not only for vessel detection in UWFFA images but also for other FA images. To demonstrate this, we also trained the alternative network architectures using (only) the UWFFA images in the RECOVERY-FA19 dataset and tested the vessel detection performance on the fundus FA images in the DRIsfahanCFnFA dataset [29]. The ground truth data, which contains 59 fundus FA images, is obtained using the cross-modality transfer described in Section 4. Quantitative results are reported in Fig. 13 and Table 3. The best performing method achieves an AUC ROC of 0.974, an AUC PR 0.887, and the maximum DC of 0.808, outperforming other baseline methods. The unsupervised methods, SFAT [13] and MSMA [1], are developed for detecting vessels from UWFFA images rather than fundus FA and thus have relatively low DC (0.607 and 0.691, respectively). Deep neural networks show significant improvement over unsupervised methods. In Fig. 12, we show visual results of the vessel detection obtained with the Prop. + GAN and the Prop. +

NestUNet. While deep neural networks are trained only using the UWFFA images from the RECOVERY-FA19 dataset, it has the generalization ability to detect vessels from fundus FA images. Compared with the Prop. + NestUNet, the GAN loss term $\mathcal{L}_{GAN}$ encourages global level consistency between the predicted vessel maps and the ground truth, as shown in red rectangles in Fig. 12.

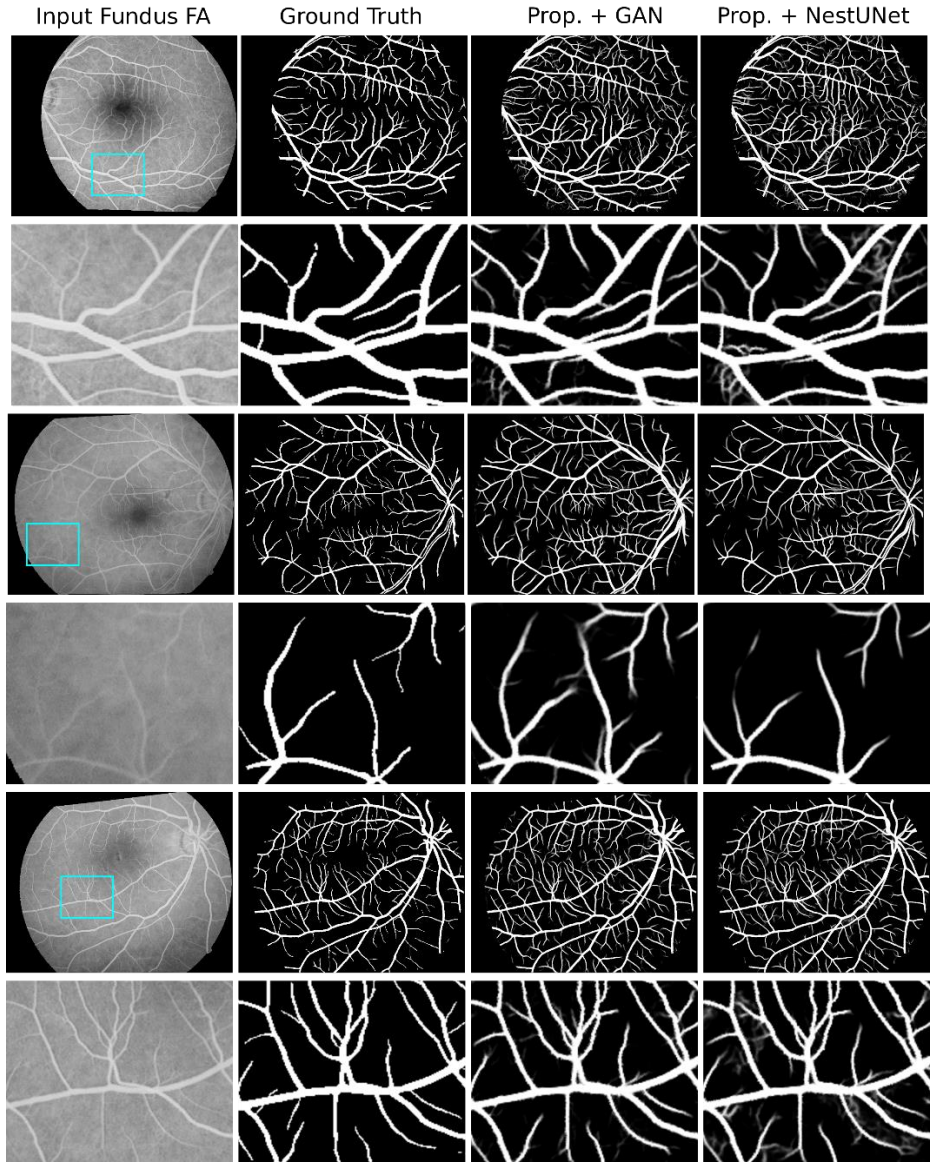

*Figure 12: Qualitative comparison of results from the DRIsfahanCFnFA dataset [29]. The rectangular regions show that the proposed adversarial network produces more accurate detections than NestUNet [42].*

| Methods | AUC ROC | AUC PR | Max DC |
|---|---|---|---|
| SFAT [13] | - | - | 0.607 |
| MSMA [1] | - | - | 0.691 |
| VDGAN [3] | 0.965 | 0.851 | 0.776 |
| Prop. + UNet | 0.972 | 0.883 | 0.802 |
| Prop. + NestUNet | 0.972 | 0.882 | 0.804 |
| Prop. + GAN | **0.974** | **0.887** | **0.808** |

Table 3: Quantitative comparison of the results obtained from different methods on the DRIsfahanCFnFA dataset. The best result is shown in bold.

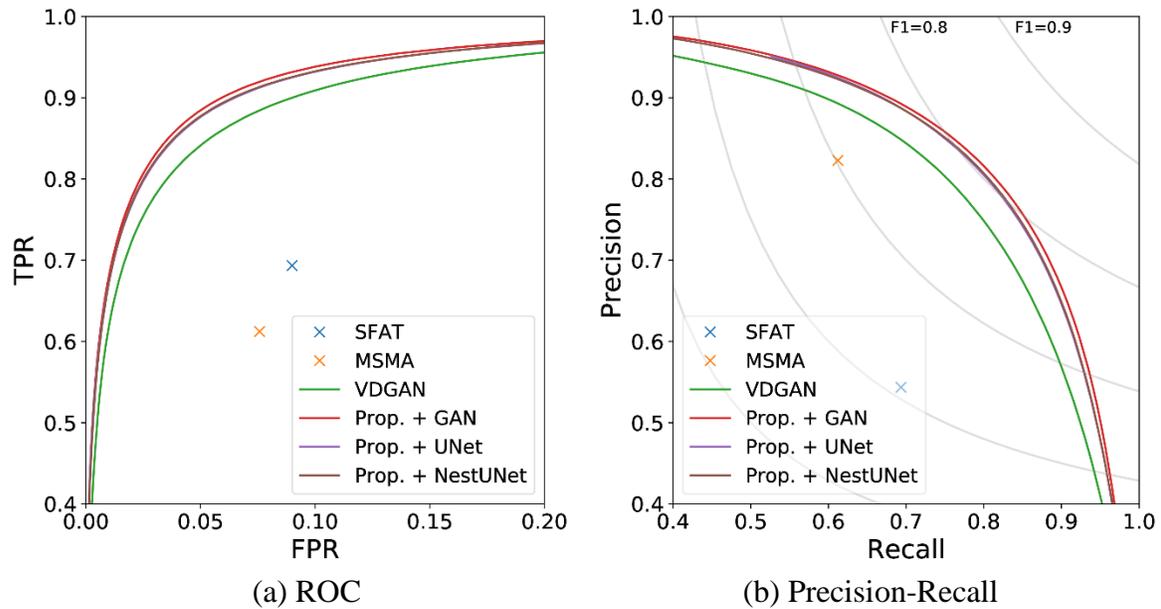

(a) ROC  (b) Precision-Recall

Figure 13: ROC and Precision-Recall curves for different methods on the DRIsfahanCFnFA dataset [29]. The gray curves represent the isolines of dice coefficients.

## 7. Conclusion

We proposed a novel deep learning pipeline for detecting retinal vessels in FA images. Using a cross-modality approach and a human-in-the-loop approach, our pipeline significantly reduces the effort required for generating labeled ground truth images. Experimental validations on three datasets, including a new RECOVERY-FA19 UWFFA dataset, demonstrate that the proposed pipeline significantly outperforms existing methods. To facilitate further development and evaluation of retinal vessel detection in FA images, the RECOVERY-FA19 dataset is made publicly available [46][4].

The proposed pipeline provided a particularly useful methodology for generating labeled ground truth data. While our focus here was on labeling vessels in FA retinal images, the key underlying ideas could be applied in other situations. The registration approach that we describe in Section 4 can also be used to facilitate identification and comparison of longitudinal vessel changes, preliminary results on which have been reported in [2]. The idea of cross-modality (label) transfer by registering observations of the same object captured with different modalities is potentially useful in speeding up other ground truth labeling tasks. Used in combination with the human-in-the-loop approach, such methods can significantly reduce tedium and improving engagement, and improve availability of datasets with accurately labeled ground truth, which is currently a key bottleneck in deploying deep learning solutions for a number of problems.

## Acknowledgments

The work was supported in part by a University of Rochester Research Award, by a distinguished researcher award from the New York state funded Rochester Center of Excellence in Data Science (contract CoE #3B C160189) at the University of Rochester, by an unrestricted grant to the Department of Ophthalmology from Research to Prevent Blindness, and grant P30EY001319-35 from the National Institutes of Health. We would like to thank the Center for Integrated Research Computing, University of Rochester, for providing access to computational resources. We also thank Shaun Lampen, Alex Rusakevich, and Brenda Zhou for collecting UWFFA images from the RECOVERY trial.

---

[4] A sample low resolution annotated image is currently provided and the full set of 8 high resolution images will be made available with the publication of the paper.

# Supplementary Material

## 1. Overview

This document provides supplementary material for the main paper. In Section S.2, we provide detailed derivations for parameter updates in the M-step for the robust EM based chamfer registration. In Section S.3, we provide the implementation details, additional network architectures, and the parameter settings. In Section S.3.3, we include additional results for the VAMPIRE dataset.

## 2. Detailed Derivation for Parameter Estimations in the M-Step

The expectation of the complete-data log likelihood $Q(\theta, \widehat{\theta}^{(l)})$ is

$$
\begin{aligned}
Q(\theta, \widehat{\theta}^{(l)}) &= \mathbb{E}[\sum_{j=1}^{N_j} \log p(d_j, w_j|\theta)] \\
&= \sum_{j=1}^{N_j} \sum_{w_j \in \{0,1\}} p(w_j|d_j, \theta) \log p(d_j, w_j|\theta) \\
&= \sum_{j=1}^{N_j} p_j^{(l)} [-\lambda d_j + \log(\pi) + \log(\lambda)] \\
&\quad + (1 - p_j^{(l)})[\log(1 - \pi) - \log(D_{max})].
\end{aligned}
\quad (S.1)
$$

Setting the derivatives of $Q(\theta, \widehat{\theta}^{(l)})$ in (S.1) with respect to $\pi$, $\lambda$, and $\beta$ to zero, we obtain the expression for the optimal parameters in (7) and (8). The optimization problem in (8) can be solved by Levenberg-Marquardt (LM) algorithm that is an iterative method for solving non-linear least squares problems. The LM algorithm starts with an initial estimate $\widehat{\beta}^{(0)}$ and proceed to refine the parameters $\beta$ at each iteration. Specifically, $\beta$ is adjusted by a parameter increment $\Delta$ to $\beta + \Delta$, where $\Delta$ is obtained by solving the linear system of equations

$$
(\sum_{j=1}^{N_j} J_j^T J_j + \sigma I)\Delta = 2 \sum_{j=1}^{N_j} p_j J_j r_j, \quad (S.2)
$$

where $r_j$ is the residual vector for point $r_j$ that can be efficiently calculated by using the distance transform, $I \in \mathbb{R}^{12 \times 12}$ is the identity matrix, $J_j \in \mathbb{R}^{2 \times 12}$ is the Jacobian matrix at each transformed target point $\mathcal{T}_\beta(q_j)$, which is computed as

$$
\frac{\partial \mathcal{T}_\beta(q_j)}{\partial \beta} = \begin{bmatrix} 1 & u_j & v_j & u_j^2 & u_j v_j & v_j^2 & 0 & 0 & 0 & 0 & 0 & 0 \\ 0 & 0 & 0 & 0 & 0 & 0 & 1 & u_j & v_j & u_j^2 & u_j v_j & v_j^2 \end{bmatrix}, \quad (S.3)
$$

and $\sigma$ is a damping parameter varying from iteration to iteration. If the increment $\Delta$ leads to a reduction in the error, then $\sigma$ is divided by a factor of 10 for the next iteration, whereas if **Δ** gives an increased error, $\sigma$ is then multiplied by the same factor. This process is repeated until the convergence criterion is met.

## 3. Implementation Details

### 3.1. Preliminary Vessel Detection in FA Image

For the preliminary vessel detection using morphological analysis, we use two scales: one at the original image resolution and the other with a down-sample rate of 2. At each scale, 9 linear structuring elements (every 20) are used in the modified top-hat operation of (1) (in the main manuscript). The lengths of structuring elements are 6 for the original scale and 3 for the down-sampled one.

### 3.2. Network Architectures and Loss Function

The Prop. + UNet uses the same network architecture as the generator in the Prop. + GAN, as shown in Fig. 6 (the generator block). For the Prop. + NestUNet, we adopt the NestUNet architecture. All convolutional layers use $3 \times 3$ kernels with stride 1. We use MaxPooling layers with $2 \times 2$ kernels to reduce spatial resolutions. The objective function in (10) (in the main manuscript) is used for the Prop. + UNet and the Prop. + NestUNet.

### 3.3. Training Protocol

We feed the network $256 \times 256$ patches extracted from the FA training data with a fixed stride length 128. Patches that contains less than 1% vessel pixels are excluded. To prevent neural networks from over-fitting, we further enlarge the training set by performing on-the-fly data augmentation, i.e., randomly applying a list of transformations with different probabilities to each image before feeding into neural network as training data. Specifically, we consider following transformations: (1) rotating the image by an angle from -90° to 90°, (2) horizontally and vertically flipping the image, (3) scaling the image by a factor of 2, (4) blurring the image using Gaussian filter, and (5) adjusting the brightness and contrast of the image.

The network parameters are optimized using Adam optimizer on a NVidia Tesla V100 GPU. The learning rate is fixed as 0.001. The coefficients used for computing running averages of gradient and its square are 0.9 and 0.999, respectively. The batch size is 16 and the training dataset is shuffled between epochs. We split the data into a training set (80%) and a validation set (20%) and use the model that has the best performance on the validation set. The lambda in (11) is set to 1.

## 3.4. Additional Evaluations on the VAMPIRE Dataset

For completeness, we also provide the quantitative metrics comparing the different methods on the VAMPIRE dataset. Table S.1 lists the AUC for both ROC and PR curves and the maximum DC using the limited accuracy original (vessel) labeling. In addition, we also report in Table S.1 results evaluated on the "AMD2" and the "GER4" images using both the original labeling and the (refined) ground truth. It is interesting to see that the accuracy of the unsupervised methods, SFAT and MSMA, is decreased when they are evaluated using the (refined) ground truth. This is because the (refined) ground truth data contains more vessel that are not detected by SFAT and MSMA (more false negative detections). The results obtained with deep neural networks, which are evaluated using the (refined) ground truth, are indeed better than the ones accessed with the original labeling.

| Methods | All Test Images: Original Lbl. | | | "AMD2" and "GER4": Original Lbl. | | | "AMD2" and "GER4": Refined GT | | |
|---|---|---|---|---|---|---|---|---|---|
| | AUC ROC | AUC PR | Max DC | AUC ROC | AUC PR | Max DC | AUC ROC | AUC PR | Max DC |
| SFAT | - | - | 0.624 | - | - | 0.647 | - | - | 0.573 |
| MSMA | - | - | 0.647 | - | - | 0.713 | - | - | 0.654 |
| VDGAN | 0.957 | 0.702 | 0.686 | 0.971 | 0.738 | 0.680 | 0.965 | 0.783 | 0.707 |
| Prop. + UNet | 0.978 | **0.786** | 0.715 | 0.987 | **0.809** | 0.713 | 0.994 | 0.948 | 0.878 |
| Prop. + NestUNet | **0.979** | 0.779 | 0.715 | 0.987 | 0.801 | 0.727 | 0.995 | **0.953** | **0.883** |
| Prop. + GAN | 0.978 | 0.780 | **0.715** | **0.988** | 0.808 | **0.731** | **0.995** | 0.950 | 0.878 |

*Table S.1: Quantitative comparison of the results obtained from different methods on the VAMPIRE dataset. All deep neural networks are trained on networks are trained on the RECOVERY-FA19 dataset. The best result is shown in bold.*